\documentclass{article}

\usepackage{PRIMEarxiv}

\usepackage[utf8]{inputenc} 
\usepackage[T1]{fontenc}    
\usepackage{hyperref}       
\usepackage{url}            
\usepackage{booktabs}       
\usepackage{amsfonts}       
\usepackage{nicefrac}       
\usepackage{microtype}      
\usepackage{lipsum}
\usepackage{fancyhdr}       
\usepackage{graphicx}       
\graphicspath{{media/}}     
\usepackage{algorithm}
\usepackage{algpseudocode}
\usepackage{subfig}
\usepackage{multirow} 
\usepackage{caption} 
\usepackage{tabularx} 
\usepackage{amsmath}

\title{CAAP: Class-Dependent Automatic Data Augmentation Based on Adaptive Policies for Time Series
}

\author{
  Tien-Yu Chang \\
  Institute of Computer Science and Engineering\\
  National Yang Ming Chiao Tung University\\
  Hsinchu\\
    \And
  Hao Dai \\
  Institute of Information Management\\
  National Yang Ming Chiao Tung University\\
  Hsinchu\\
   \AND
   Vincent S. Tseng* \\
   Department of Computer Science \\
   National Yang Ming Chiao Tung University, Hsinchu, Taiwan, R.O.C \\
   orcid=0000-0002-4853-1594 \\
   \texttt{vtseng@cs.nycu.edu.tw} \\
}

\begin{document}
\maketitle

\begin{abstract}

Data Augmentation is a common technique used to enhance the performance of deep learning models by expanding the training dataset. Automatic Data Augmentation (ADA) methods are getting popular because of their capacity to generate policies for various datasets. However, existing ADA methods primarily focused on overall performance improvement, neglecting the problem of \textit{class-dependent bias} that leads to performance reduction in specific classes. This bias poses significant challenges when deploying models in real-world applications. Furthermore, ADA for time series remains an underexplored domain, highlighting the need for advancements in this field. In particular, applying ADA techniques to vital signals like an electrocardiogram (ECG) is a compelling example due to its potential in medical domains such as heart disease diagnostics.

We propose a novel deep learning-based approach called \textit{Class-dependent Automatic Adaptive Policies (CAAP)} framework to overcome the notable class-dependent bias problem while maintaining the overall improvement in time-series data augmentation. Specifically, we utilize the policy network to generate effective sample-wise policies with balanced difficulty through class and feature information extraction. Second, we design the augmentation probability regulation method to minimize class-dependent bias. Third, we introduce the information region concepts into the ADA framework to preserve essential regions in the sample. Through a series of experiments on real-world ECG datasets, we demonstrate that CAAP outperforms representative methods in achieving lower class-dependent bias combined with superior overall performance. These results highlight the reliability of CAAP as a promising ADA method for time series modeling that fits for the demands of real-world applications.

\end{abstract}


\section{Introduction}\label{introduction}
\textit{Data Augmentation (DA)} is a promising means to extend the training dataset with new artificial data to enhance the generalization capability of deep learning models. Over the years, various data augmentation methods and transformations (e.g., Gaussian noise, Cutout and Mixup \cite{zhang2017mixup}) have been proposed to extend the training dataset and avoid overfitting. Recently, \textit{Automatic Data Augmentation} (ADA) \cite{AA, randaug, fasterAA, dada} has become more popular because it can adaptively learn augmentation policies according to the dataset.

However, current ADA methods are designed to enhance overall performance, which averages performance over all classes and overlooks class performance differences. These methods introduce significant biases toward specific classes. In particular, applying data augmentation methods may introduce an unknown performance loss for particular classes, referred to as \textit{class-dependent bias} \cite{DAeffects}. Notably, this bias can persist even when data augmentation improves overall performance, rendering it unreliable for practical applications. For instance, this issue may cause the model to miss patients with rare diseases, delaying timely intervention. 

\begin{figure*}[htbp]
\centerline{\includegraphics[width=140mm,scale=0.8]{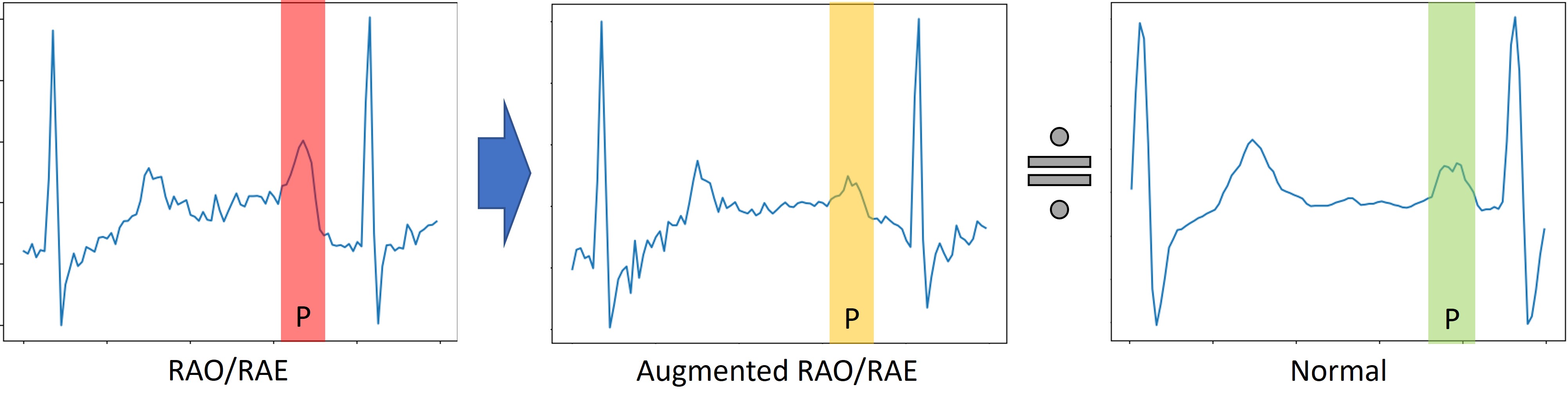}}
\vspace{-0.4cm}
\caption{Class-dependent Bias in ECG Task.
The right graph is the RAO/RAE ECG signal, the medium graph represents the augmented RAO/RAE ECG signal, and the right chart is the normal ECG signal.
Also, the red, yellow, and green boxs are the P-wave parts of the RAO/RAE, augmented, and normal ECG signals.
We use the scaling transformation to augment the ECG signal, which multiplies each time step of the original signal with factors from a normal distribution(mean=1; standard deviation=0.3).
}

\label{figintropic}
\end{figure*}

To illustrate, consider the case of an ECG task. Scaling the heartbeat may not affect the rhythm abnormally but may disrupt the waveform diagnostic. Specifically, the RAO/RAE occurs when the right atrium is larger than usual, resulting in higher P waves. On the other hand, the LAO/LAE appears when the right atrium is larger than normal, leading to longer P waves \cite{rae_ecg}. As a result, the 'Scaling' transformation may impact more on the diagnosis of RAO/RAE but less on LAO/LAE, as it alters the height of P waves. As shown in Figure \ref{figintropic}, the abnormal magnitude of P-wave in the RAO/RAE ECG signal becomes similar to the normal ECG signal after the 'Scaling' transformation.

This work aims to reduce the class-dependent bias in ADA while improving overall performance in supervised learning-based classification problems.

\textbf{Challenge 1:} Class-wise or sample-wise ADA \cite{adaaug, cadda} that adaptive implement augmentation for each class is introduced to reduce the class-dependent bias while expecting to maintain the overall performance. However, due to the extremely large search space and the limitation of the existing frameworks, their adaptive policy does not benefit the model or even degrade the model's overall performance. Therefore, identifying robust optimization targets and establishing efficient class-specific augmentation search methods are essential to address the improvement of overall performance through sample-wise ADA.

\textbf{Challenge 2:} The relationship between model class-dependent bias and overall performance may involve trade-offs. With the reasonable assumption that data augmentation introduces unknown biases for specific classes, these biases could render the model unreliable for certain data subsets (classes). Conversely, not using any data augmentation eliminates bias but offers no benefits for overall performance. This dilemma makes designing a sample-wise policy for both overall performance and class-dependent bias a challenging task. Consequently, it is both crucial and difficult to determine the relationship between class-dependent bias and overall performance, as well as to address this trade-off issue. In addition, efficient evaluation metrics for class-dependent bias have been lacking.

\textbf{Challenge 3:} Most ADA researches aim to learn the optimal magnitude (m) and probability (p) of each transformation. However, the information region, like an object's location or texture in an image or PQRST waves in ECG signals, is another crucial factor that needs to be considered. Not considering informative regions in the data sample during augmentation transformation may cause bias in the augmented sample and damage the model's performance. For instance, cutting out the object on images or masking abnormal waves in ECG signals will cause the model to learn on biased augmented samples. Moreover, different datasets or classes may have different information regions. Therefore, designing an adaptive method to learn the class/sample-specific important region in samples is crucial but not well-addressed.

Besides, we observe that the studies on ADA for time series are relatively scarce compared to the image domain, even though time series data is widely used in various real-world applications, including natural language processing (NLP) \cite{dnd}, traffic prediction \cite{traffic_app} and medical time series forecasting \cite{medical_app}. This necessitates the imperative for research in this field. Within the realm of time series, Electrocardiogram (ECG) relates to the cardiology system. It is helpful to diagnose heart problems \cite{ecg_app}, which is a valuable topic in the time series field. In the later parts of this work, we will focus on ECG tasks and ADA methods for a more in-depth introduction.

To address the aforementioned challenges, we propose a novel \textit{Class-dependent Automatic Adaptive Policies (CAAP)} framework.
The CAAP framework aims to tackle the class-dependent bias problem in ADA while improving overall performance for supervised learning-based time series classification. 
Specifically, the CAAP framework consists of three key modules: \textit{Class Adaption Policy Network}, \textit{Information Region Adaption} and \textit{Class-dependent Regulation} module.
To overcome \textbf{challenge 1}, the Class Adaption Policy Network is designed to learn augmentation policy for training sample's difficulty and similarity based on its feature and label information. With the additional label information and proper augmentation target, this module can create an efficient class-wise augmentation policy for training samples.
For \textbf{challenge 2}, the Class-dependent Regulation module handles the probability while applying augmentation policies to reduce class-dependent bias as we illustrate that there exists a trade-off relation between accuracy and class-dependent bias, as well as the best performance augmentation percentage positively relative to class-wise performance. Hence, a no-augmentation reweighting process is adopted within this module to regulate the class-dependent bias further. Also, we measure the class-dependent bias from data augmentation based on the confusion matrix.
For \textbf{challenge 3}, the Information Region Adaption module aims to keep the critical part of the sample and create more reasonable and practical augmentation samples. The saliency map method is implemented to find and protect information regions based on the sample's class and feature.

Our contributions are three-fold: 1) We propose a CAAP framework, the first attempt to regulate class-dependent bias and improve overall performance by class-adaptive policies reweighting and information region preserving. 2) We introduce a new metric to measure the class-dependent bias in data augmentation and demonstrate a trade-off between accuracy and class-dependent bias reduction. 3) We introduce additional class-wise and sample-wise information in ADA policies to create a low-bias and effective augmentation policy.

The remainder of this work is organized as follows: Section \ref{relatedwork} introduces the related work. The problem definition and the proposed methods are described in Section \ref{method}. Later in Section \ref{experiment_setting} and Section \ref{experiment}, we perform a series of experiments to evaluate our proposed method on real-world datasets. Finally, Section \ref{conclusion} discusses and summarizes this work's contributions and future work.

\section{Related Work}\label{relatedwork}

\subsection{Data Augmentation and Automatic Data Augmentation}

Data augmentation methods can be divided into four common approaches \cite{survey1, survey2}.
Transform-based and mixing-based methods \cite{SMOTE, zhang2017mixup} create new samples through domain knowledge or mixing and recombining existing samples.
Some of them consider important regions in samples \cite{saliencymix, keepaug, gradcam}.
Generative-based methods like \textit{Generative Adversarial Networks (GAN)} \cite{GAN} use generative models to produce synthetic samples.
Automatic Data Augmentation (ADA) methods collect a set of transformations and learn a policy to apply them to specific datasets, including Auto-Augment \cite{AA}, PBA \cite{PBA} and MODALS. However, it takes a lot of time to search for the policy.
To reduce the search space, RandAugment \cite{randaug} and TrivalAugment \cite{trivalaug} use a uniform distribution as their policy, while Faster AA \cite{fasterAA} and DADA \cite{dada} learn the policy through gradient descent.
Some ADA studies focused on finding augmented samples with affinity and diversity \cite{diffsim}, including \textit{AugMax} \cite{augmax}, TeachAugment \cite{teachaug} and DND \cite{dnd}.

Recent studies have proposed more adaptive ways for ADA, including CADDA \cite{cadda}, which learns augmentation policies for each class, and AdaAug \cite{adaaug}, which utilizes a policy sub-network to learn a policy for each sample. They used class-balanced accuracy or accuracy of the test dataset to evaluate their methods.
However, their adaptive policy has limited benefits due to the extreme search space or limitations of their policy optimization.
Additionally, Randall Balestriero et al. \cite{DAeffects} proposed a study about how applying data augmentations can introduce an unknown bias of specific classes (class-dependent bias). The study uses detailed experiments to demonstrate that standard data augmentation introduces class-dependent bias, which rapidly reduces the accuracy of some class when augmentation strengthens.
While some studies have investigated class-wise ADA and bias from data augmentation, the relationship between data augmentation and bias in different classes still needs further exploration. Also, previous studies used time-costuming experiments to show the class-dependent bias from data augmentation. There's a lack of efficient evaluation metrics for class-dependent bias. Furthermore, existing ADA methods have prioritized overall performance at the expense of performance in specific classes, leading to a class-dependent bias for these classes.


\subsection{Data Augmentation for ECG signal}

Most data augmentation papers for ECG signals were transform-based methods that heavily rely on human setting parameters or well-annotated ECG signals. Only a few considered GAN or ADA methods for data augmentation.
Most ECG Augmentation studies utilize the PQRST waveform property to design their methods \cite{QRS, RR, ECGAug, ecg_augment}.
Some of the ECG Augmentation methods \cite{ecg_randaug, taskaug} combine ECG-specific noises and SOTA ADA studies.
Although various studies are about Data Augmentation for ECG signals, they were limited to using ECG transformations or representative ADA methods without considering introduced bias for specific classes or ECG region properties.

\section{Proposed Method}
\label{method}

\subsection{Problem Definition}

Automatic Data Augmentation (ADA) methods aim to automatically search for improved augmentation policies on a transformation set for the specific dataset. These policies determine the appropriate transformations to apply during model training, aiming to minimize validation loss and enhance generalization performance.
Formally, Let $TS$ be a transform set where each $\tau_j$ represents the j-th operation in the transformation set.
The policy parameters of ADA methods, denoted as $P$, include probabilities $p$ and magnitudes $m$ for each transformation $\tau$ in $TS$.
The policy parameter can be formulated as $P = {(p_j,m_j), 1 \leq j \leq |TS|}$, where $|TS|$ represents the size of the transformation set.
For a data sample $x \in D$, ADA methods select transformation $\tau_j$ (including the option of no augmentation) with probability $p_j$ and apply it to the sample with magnitude $m_j$ as $x \mapsto \tau_j(x,m_j)$.
This mapping is repeated $K$ times, typically determined by the user.
The sequence of transformations, guided by the policy parameter $P$, is denoted as $T_P(x) = \tau^K(x,m) \circ \dots \circ \tau^1(x,m) = x'$, which $T_P$ represents a series of transformations $\tau$ and $x'$ is augmented data sample.
With train dataset $D_{train}$ and validation dataset $D_{valid}$, the goal of ADA methods is to solve the following bi-level optimization problem:
\begin{equation}
min_{T_P}L(\theta^* | D_{valid}) s.t. \theta^* \in arg min_{\theta}L(\theta | D'_{train})
\end{equation}where $\theta$ represents the parameters of predictive model, $L$ is the loss function and $D'_{train}=T_P(D_{train})$.

In this work, our objective is to learn a policy network $h$ that takes the feature embedding $f(x_i)$ and labels $y_i$ of a sample $x_i$ as input and learns class-wise policy parameters $P_{i}$ as $h_w(f(x_i),y_i) = P_{i}$, where $(x_i,y_i) = D_{train}$.
We aim to find the optimal parameter $w$ for the policy network, which minimizes the validation loss and maximizes the augmented training loss when searching for the best model parameters using the augmented training dataset.

The optimization problem can be formulated as follows:
\begin{equation}
\begin{aligned}
min_{w}(L(\theta^* | D_{valid})+L(\theta^* | D'_{train})^{-1}) s.t. \theta^* \in arg min_{\theta}L(\theta | D'_{train})
\end{aligned}
\end{equation}where $h_w$ represents the policy network with optimal parameter $w$, augmented training dataset represents as $D'_{train}=T_P(D_{train})$ and $P=h_w(D_{train})=h_w(f(x_i),y_i)(x_i)$.

\subsection{Proposed Framework}

\begin{figure*}[htbp]
\centerline{\includegraphics[width=140mm,scale=0.8]{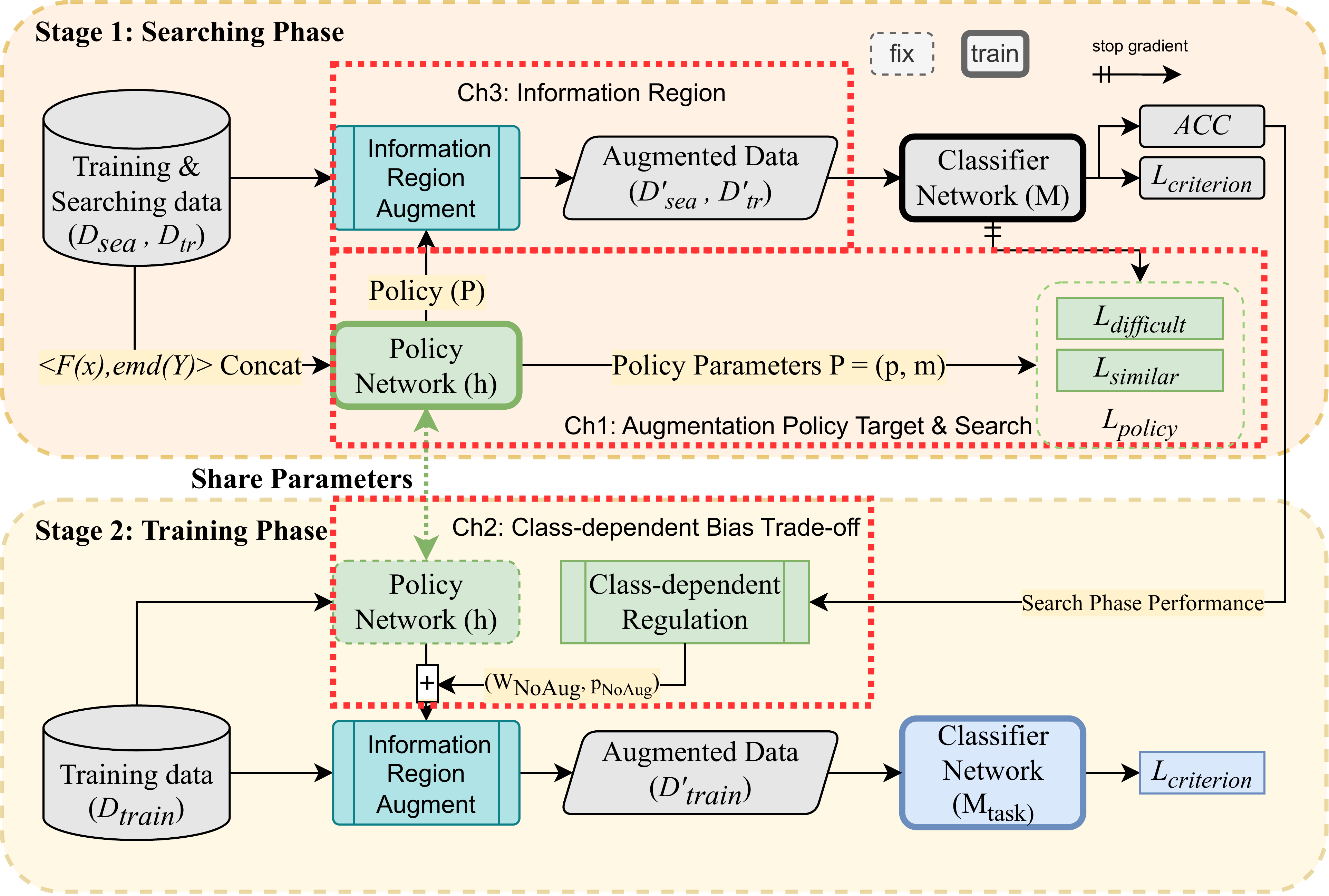}}
\vspace{-0.4cm}
\caption{CAAP Framework Overview.
(Stage 1) Searching Phase: Search for augmentation policy and other parameters;
(Stage 2) Training Phase: Using policy to train the classifier network with the Class-dependent Regulation module.
}
\label{figframework}
\end{figure*}

The proposed CAAP framework, illustrated in Figure \ref{figframework}, comprises three key components, which dual with three mentioned challenges, shown as three red dotted blocks.
First, the Class Adaption Policy Network optimizes the policy loss of search data to learn an augmentation policy that captures the relationship between sample information and suitable augmentation policy.
Second, the Class-dependent Regulation module adjusts the weight of the no-augmentation transformation (NoAug) based on the performance of each class during the search phase, effectively addressing class-dependent bias while training the task model.
Finally, the Information Region Adaption module preserves essential information in the data sample throughout both phases, effectively addressing the challenge of maintaining informative regions during augmentation.

The dataset for searching phase $D_{search}$ is divided into two subsets: training subset $D_{tr}$ and searching subset $D_{sea}$. On the other hand, the dataset for training phase $D_{train}$ uses all data to train the framework.
In the searching phase, the framework is trained by iteratively updating the model $M$ and policy network $h$. The model $M$ only uses augmented training subset $D'_{tr}$ for training, and the policy network $h$ uses both subsets for training through our policy loss, $L_{policy}$.
During the training phase, the policy network, in conjunction with the Class-dependent Regulation module, is utilized for training the task model $M_{task}$ through the whole dataset $D_{train}$.
The module adjusts the policies $P$ obtained from the policy network to mitigate the class-dependent bias.

\subsection{Class Adaption Policy Network}

The Class Adaption Policy Network aims to reduces class-dependent bias through additional sample embedding and labeling information and introduce a policy network to search for augmentation policy efficiently. With the help of this module, the CAAP framework can learn optimal augmentation policies for each class or sample, which reduce the bias of every class.
Figure \ref{figframework} illustrates our policy network, denoted as $h$, which is a simple neural network that takes the feature embedding $F(x)$ and labels information $Y$ of the data sample as input and outputs the policy parameters $P$ for data augmentation.

The feature embedding is obtained from the sample's representation after the final pooling layer. At the same time, the label information is transformed into one-hot encoding and passed through the label embedding layer $emd()$ to create the label embedding. The policy network concatenates the feature and label embeddings to calculate the augmentation policy. The output consists of weight ($p$) and magnitude ($m$) parameters, denoted as $h(F(x), emd(Y)) = (p', m')$. The weight parameter $p$ represents the probability of selecting a transformation, while the magnitude parameter $m$ indicates the strength of the transformation. We apply the softmax function $soft()$ to output weight logits, ensuring that the sum of weights for all transformations is one. The sigmoid function $\sigma$ is applied to output magnitude values, constraining its value between 0 and 1. Thus, the final policy parameters are formulated as $P=(p, m)=(soft(p'), \sigma(m'))$.

To optimize the policy network, we employ the policy loss and adopt techniques from Gumbel-softmax \cite{dada} tricks and AdaAug \cite{adaaug}. In the searching step, we augment the original sample using all transformations in the transformation set and append magnitude parameters using Gumbel-softmax tricks. The group of augmented samples with all transformations are denoted as $\hat{x}$. The feature extractor of the model is then used to obtain the feature embeddings of the augmented samples. We calculate the weighted sum of feature embeddings $F'(\hat{x})$ with weights of each transformation. The formula is defined as:

\begin{equation}
F'(\hat{x}) = \frac{1}{|TS|} (\sum_{j=1}^{|TS|}p_jF(\tau_j(x)))
\end{equation}where $TS$ is the transformation set, $\hat{x}$ is augmented samples with all transformation, $p_j$ is the probability of transformation with index $j$ and $F(\tau_j(x))$ represent the feature embedding of augmented samples $\tau_j(x)$.

Finally, we utilize the classifier $g$ of the model $M$ to classify the sum of embeddings and perform gradient descent on the policy loss ($L_{policy}$). Our proposed policy loss consists of two components: difficult loss ($L_{difficult}$) and similar loss ($L_{similar}$). The formula is as follows:
\begin{equation}
L_{policy} = \frac{L_{difficult}(D_{tr}) + L_{similar}(D_{sea})}{2}
\end{equation}

Inspired by recent research \cite{diffsim, dnd, augmax}, we design the difficult and similar losses to ensure the diversity and affinity of our augmented samples. The difficult loss aims to increase the diversity of the learned samples, forcing the model to learn more from confident samples. The similar loss aims to ensure that the augmentation policy becomes similar to the original sample for unseen samples, controlling the difficulty. The formulas for the losses are as follows:

\begin{equation}
L_{difficult}(D_{tr}) = \frac{w(x,\hat{x},y)L_{train}(x,y)}{L_{train}(\hat{x},y)}, (x,y) \in D_{tr}
\end{equation}where $L_{train}(x,y)$ represents the loss of original training branch from $D_{tr}$, $L_{train}(\hat{x},y)$ is the loss of same training branch with augmentation and $w(x,\hat{x},y)$ is a weighting factor of difficult loss.
\begin{equation}
w(x,\hat{x},y) = \sqrt{p_y(x)(max\{p_y(x)-p_y(\hat{x}),0\})}
\end{equation}where $p_y(x)$ and $p_y(\hat{x})$ represent confidence of origin and augmented data.

\begin{equation}
L_{similar}(D_{sea}) = \frac{L_{search}(\hat{x},y)}{L_{search}(x,y)}, (x,y) \in D_{sea}
\end{equation}where $L_{search}(x,y)$ represents the loss of original searching branch from $D_{sea}$ and $L_{search}(\hat{x},y)$ is the loss of same searching branch with augmentation.

We normalize our difficult and similar losses by the losses of the original sample to make both losses equally important. In the difficult loss, we further emphasize the difficulty of high-confidence training data by confidence reweight. With the policy network and loss, our framework can generate diverse and similar class-dependent augmentation samples based on feature and label information.
For the training phase, we utilize the Class Adaption Policy Network to generate augmentation policies for each sample and train the classifier network ($M_{task}$) through classification loss ($L_{criterion}$). In this work, we use cross-entropy loss as our classification loss.

\subsection{Class-dependent Regulation}

The Class-dependent Regulation module aims to mitigate class-dependent bias effectively while applying data augmentation. Based on the experimental observations, we propose a systematic approach that accounts for the performance of each class. By adjusting the weight of no-augmentation, we effectively regulate the class-dependent bias while preserving overall performance.

As depicted in Figure \ref{figframework},
this module incorporates an additional no-augmentation weight, denoted as $p_{NoAug}$, to the probability parameters $p$ to obtain less biased policies for each class during the training model.
To determine the no-augmentation weight, we utilize class-wise recall in the searching phase to measure class performance. The weight is computed as follows:
\begin{equation}
W_{NoAug} = \alpha_{NoAug} \times (1.0 - recall_{search})
\end{equation}where $\alpha_{NoAug}$ represents the strength of regulation, $recall_{search}=\{recall_{search}^{c}, 1 \leq c \leq N_C\}$ is class-wise recall and $N_C$ is total class number.
The no-augmentation weight is then added to the policy parameter using the following equation:
\begin{equation}
p_{new} = (1.0 - W_{NoAug}) \times p_{old} + W_{NoAug} \times p_{NoAug}
\end{equation}where $p_{old}$ denotes the possibility parameter from the policy network, $p_{NoAug}$ represents the possibility parameter from the policy that does not employ any data augmentation and $p_{new}$ is the updated possibility parameter.

By reweighting the original augmentation policy, our method generates more reliable policies for each class and effectively regulates class-dependent bias during training. This approach considers class performance in the search phase, enabling the reduction of bias while preserving overall performance.

\subsection{Information Region Adaption}
\begin{figure}[htbp]
\centerline{\includegraphics[width=80mm,scale=1.0]{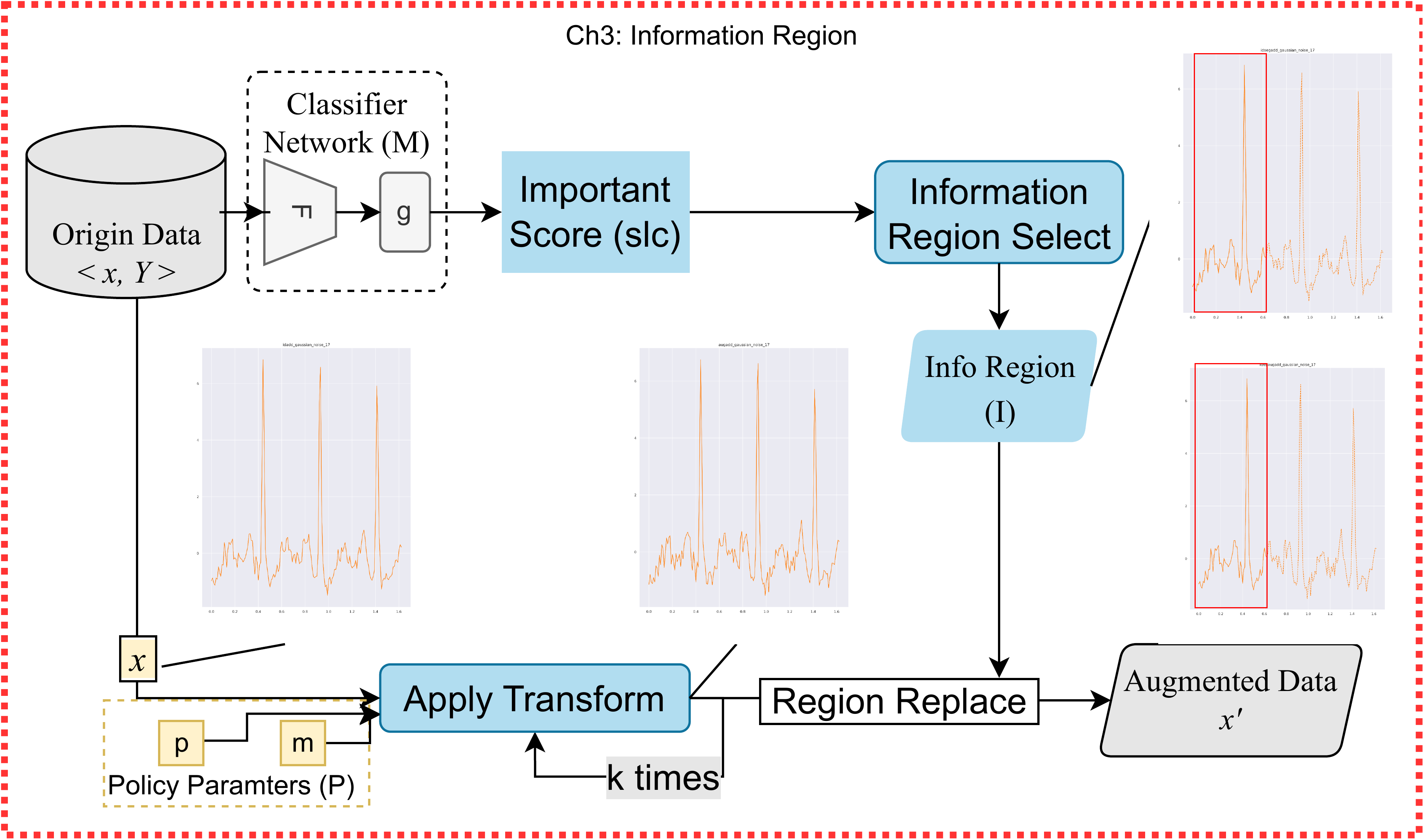}}
\caption{Diagram of Information Region Adaption Module.}%
{
In the upper branch, the origin sample $x$ is passed into the classifier network to calculate importance scores $slc$, then randomly select a region $I$ from the origin sample with higher importance scores.
In the lower branch, the module augments the origin sample $x$ $k$ times based on policy parameters ($p$ and $m$). After that, the selected information region $I$ will paste back to the augmented sample, generating the final augmented sample $x'$.
}
\label{figinforeg}
\end{figure}

The information region, such as PQRST wave properties in ECG signals, is a critical concept of transform-based data augmentation \cite{RR, ECGAug, keepaug}.
However, current time series ADA studies ignore these crucial regions in signals during augmentation transformation, which causes the class or sample-dependent bias in time series data augmentation and damages the model's performance.
Motivated by this concept, we propose our Information Region Adaptation module to detect and preserve the information region of a time series signal in augmentation transformation without modifying the actual transformation.
Figure \ref{figinforeg} illustrates the module, where the origin sample $x$ is fed into the classifier network to calculate importance scores through saliency map as $slc()$.
The saliency map \cite{saliencymap} is computed by taking the absolute value of the gradient of each time step in the sample, according to the following formula:
\begin{equation}
slc(x,y) = |\nabla_{x}l_{y}(x)|
\end{equation}where $l_{y}$ represents the loss function with respect to label $y$.

By performing average pooling on the saliency map, we obtain importance scores for all possible regions, denoted as $I(x,y,fl) = avgpool(slc(x,y),fl)$, where $avgpool()$ represents the average polling function and $fl$ is the filter length.
We then randomly select a region with average importance score higher than a given importance threshold $thres$, finding region $I(x,y,fl) \geq thres$.
The importance threshold $thres$ is a parameter between 0 and 100, indicating the percentile of all possible region importance scores $I(x,y,fl)$.
After applying data augmentation transformations to the sample, the information region adaption module pastes the informative region in the original sample back into the augmented sample to maintain critical information about the time series signal.

Our information region adaption module synergizes with our policy network during the searching phase and protects the original information during the training phase.
Consequently, it dynamically preserves informative regions and generates augmentation samples that are more reasonable and practical.
Furthermore, by leveraging the sample's feature and label information to compute the saliency map, our module can protect informative regions based on sample and class information.

\section{Experiments Settings}\label{experiment_setting}
This section illustrates the setting we used in our experiments:
First, we describe the dataset, data pre-processing and evaluation metrics for experiments.
Next, we introduce the backbone models, hyper-parameter settings and the augmentation transformations we used.
Finally, we describe the representative ECG Augmentation and ADA methods for comparison.

\subsection{Environment and Dataset}

We conduct experiments with 12-lead ECG datasets, including \textit{PTBXL} \cite{ptbxl}, \textit{Chapman} \cite{chapman} and \textit{CPSC} \cite{cpsc:icbeb}.
PTBXL is the most variant and largest ECG dataset recently proposed, which contains diagnostic, rhythm and form statements of ECG signal.
Chapman is a multi-label ECG dataset from Chapman University and Shaoxing People's Hospital, which contain rhythm and diagnostic labels.
CPSC is a varied-length ECG dataset with nine labels; sample lengths range from 6 to 60 seconds.
These datasets are representative datasets for ECG studies \cite{ecg_randaug, taskaug, chapman_used} due to their size and diversity.
 
After collecting datasets, we perform label and feature preprocessing to fit the later usage.
For PTBXL, we divide the labels into three main groups due to their electrocardiography diagnostic descriptions, including diagnostic of cardiovascular disease, form and rhythm statements.
We further cluster diagnostic labels into three levels of groups from the dataset statement:
\textit{diagnostic (diag)}, \textit{diagnostic sub-class (sub)} and \textit{diagnostic super-class (sup)}.
We filter out the labels that appear less than 300 times for the Chapman dataset to remove the extremely minor labels.
Because of the limited size of records and labels for the CPSC dataset, we did not perform any down sample method in the CPSC dataset.
Finally, for all datasets we used, we selected only single-label samples and turned the dataset into a multi-class classification dataset because most of the representative methods were designed for multi-class classification.

Feature preprocessing involves normalizing ECG signals through z-score normalization from \textit{PTB benchmark's} \cite{PTBbenchmark}. For the CPSC dataset, we first normalized raw samples because of their dynamic length, then padded them into max size (60 seconds).
We chose the 100 Hz ECG signal for PTBXL and CPSC datasets, and we down-sampled the ECG signal to 100 Hz with a scipy signal package for the Chapman dataset.
The new statistic outcome after prepossessing is shown in Table \ref{tab_data_p} and includes the number of records (Records), the number of classes (Class nums), the sampling rate (Sfreq (Hz)) and the time length (Length).

\begin{table}[htbp]
\caption{The characteristics of the ECG datasets after pre-processing.}
\begin{center}
\resizebox{0.45\linewidth}{40pt}{
\begin{tabular}{|c|c|c|c|c|}
\hline
\textbf{Dataset} & \textbf{Records}& \textbf{Class nums}& \textbf{Sfreq (Hz)}& \textbf{Length} \\
\midrule
\textbf{PTBXL diag} & 17073 & 44 & 100 & 10 s \\
\textbf{PTBXL sub} & 17261 & 23 & 100 & 10 s \\
\textbf{PTBXL sup} & 18243 & 5 & 100 & 10 s \\
\textbf{PTBXL rhythm} & 20985 & 12 & 100 & 10 s \\
\textbf{PTBXL form} & 7962 & 19 & 100 & 10 s \\
\hline
\textbf{Chapman} & 6187 & 12 & 100 & 10 s \\
\hline
\textbf{CPSC} & 6401 & 9 & 100 & 60s \\
\hline
\end{tabular}}
\label{tab_data_p}
\end{center}
\end{table}

\subsection{Evaluation Metrics}

We evaluate the overall performance of the representative and our method using the accuracy (Accuracy) and macro recall (Macro Recall), which are commonly used in data augmentation research.
As discussed previously, there is a lack of suitable metrics for measuring class-dependent bias. Therefore, we propose the development of metrics by calculating the number of error samples after applying data augmentation for each class. For instance, in medical disease classification, we aim to assess the number of patients correctly diagnosed by the trained model with and without data augmentation. By considering the correctness before and after data augmentation, we categorize the data samples into four groups using a "confusion matrix".

\begin{table}[htbp]
\begin{center}
\resizebox{0.45\linewidth}{20pt}{
\begin{tabular}{|c|c|c|}
\hline
\textbf{For class c} & \multicolumn{2}{|c|}{\textbf{Without DA}} \\
\hline
\textbf{With DA} & \textbf{Correct} & \textbf{Error} \\
\hline
\textbf{Correct} & Both Correct (STP) & DA Improve (SFP) \\
\hline
\textbf{Error} & DA Bias (SFN) & Both Error (STN) \\
\hline
\end{tabular}}
\end{center}
\end{table}

We calculate sample-wise improvement ($Swise_{improve}$) and bias ($Swise_{bias}$) of each class, and their formula are as follow:
\begin{equation}
Swise_{improve} = \frac{SFP}{(STP+SFP+SFN+STN)}
\end{equation}
\begin{equation}
Swise_{bias} = \frac{SFN}{(STP+SFP+SFN+STN)}
\end{equation}

These metrics indicate the percentage of error samples become correct and correct samples become error after applying data augmentation.
The macro average between classes is employed to evaluate both metrics.
Furthermore, we calculate sample-wise gain ($Swise_{gain}$) as $Swise_{improve} - Swise_{bias}$ to evaluate the total improvement of data augmentation in a sample-wise view.
Our objective is to maximize sample-wise gain while applying data augmentation.


\subsection{Backbone \& Hyper-parameter \& Augmentation Transformations}

We use 1d-Resnet \cite{FCN}, FCN \cite{FCN} and LSTM \cite{LSTM} as backbone models, following the PTBXL benchmark \cite{PTBbenchmark} settings.
We set the cross-entropy loss as our classification loss ($L_{criterion}$) for training the model.
Also, the batch size is set to 128, with a learning rate of 1e-2, weight decay of 1e-2, and AdamW optimizer with OneCycleLR scheduler, trained for 50 epochs.
We perform 10-fold cross-validation for each dataset and report the average performance.
For the fold division, we follow the dataset providers in PTBXL \cite{ptbxl} and CPSC \cite{cpsc:icbeb} datasets, and divide the dataset into 10-fold with the sklearn module in the Chapman \cite{chapman} dataset.

For Automatic Data Augmentation, except for MODALS \cite{modals}, we use a set of 10 transformations, including time reverse, fft surrogate, channel dropout, channel shuffle, time mask, gaussian noise, random bandstop, sign flip, frequency shift and Identity, from CADDA \cite{cadda} by removing EEG-specific transformations like rotation and channel symmetry.
For our proposed CAAP, we set the strength of regulation $\alpha_{NoAug}$ as 0.5, the filter length $fl$ as 100 (one second in ECG signal) and the importance threshold $thres$ as 60.

\subsection{Compared Baseline and Representative Methods}

The baseline method is the model without data augmentation (NOAUG).
In this work, the representative methods we selected are as follows:
\begin{itemize}
\item Data Augmentation for ECG signals:
    \begin{itemize}
    \item RR Permutation \cite{RR}: This method finds R-peaks for each sample and then permutates RR segments.
    \item QRS Resample \cite{QRS}: This method finds the signal region, duplicates and resamples them.
    \end{itemize}
\item General Automatic Data Augmentation:
    \begin{itemize}
    \item RandAugment \cite{randaug}: RandAugment introduces a vastly simplified search space and uses the grid search to find the best parameters.
    \item MODALS \cite{modals}: MODAL incorporates hard feature space augmentation, Population-Based Training (PBT) and extra loss to search the policy.
    \end{itemize}
\item Class-wise Automatic Data Augmentation:
    \begin{itemize}
    \item CADDA \cite{cadda}: CADDA first considers class-wise data augmentation and a relaxing gradient policy search to overcome the severe search space problem.
    \item AdaAug \cite{adaaug}: AdaAug uses the policy subnetwork to select augmentation policies for each instance.
    \end{itemize}
\end{itemize}
To our best knowledge, we search the best hyper-parameter settings for baseline and representative methods.
All the settings are shown below.
For RR Permutation \cite{RR} and QRS Resample, we use two average and Pan Tompkins detector from the ecgdetectors python package to detect R-peaks in ECG signal.
For RandAugment \cite{randaug}, we use grid search to find the best magnitudes ($m$) and the number ($n$) transformation to use.
For MODALS \cite{modals} and CADDA \cite{cadda}, we follow their hyperparameter setting in their study because both of them can directly use in time series datasets, which search for 50 search epochs ($step$). Also, the ADDA method is the non-class-wise version of CADDA \cite{cadda}.
For AdaAug \cite{adaaug} and our proposed method, we search the policy network learning rate from $plr=\{0.001,0.0005,0.0001,0.00005\}$ from the 10th fold validation result. We set the search frequency ($freq_{sea}$) as ten epochs.
Additionally, we set the dataset for the searching phase equal to the dataset for the training phase $D_{search}=D_{train}$. We equally divide the origin training dataset into training $D_{tr}$ and searching subset $D_{sea}$ for AdaAug and our proposed method in the searching phase.
During the training phase, we follow AdaAug settings for temperature $Temp = 3$, magnitude perturbation $\delta = 0.3$ and search for the number of operators $n \in \{1, 2, 3, 4\}$ in task model training.

\section{Experimental Result}\label{experiment}

Our experiments can be divided into four parts:
1) Class-dependent Trade-off Experiments: We investigate the trade-off between accuracy and class-dependent bias while changing augmentation probability.
2) External Evaluations: We evaluate the performance of the representative and our proposed methods using overall metrics and our proposed sample-wise metrics.
3) Evaluations in Different Backbones: We study the performance changes of competitive representatives and our proposed methods in different backbones.
4) Internal Evaluations: We perform internal evaluations to demonstrate the effectiveness of proposed modules in the CAAP framework.

\subsection{Class bias Trade-off Experiment}

\begin{figure}[htbp]
\centerline{\includegraphics[width=80mm,scale=1.0]{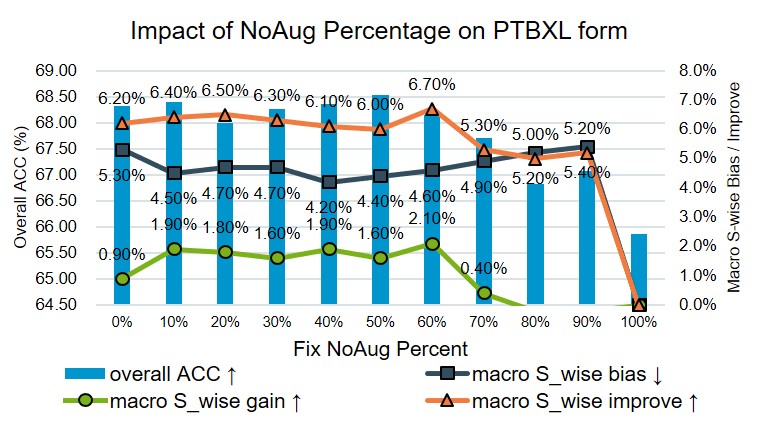}}
\vspace{-0.4cm}
\caption{Performance changes in different NoAug percentages. 
The x-axis is fixed no-augmentation percentage (\%), orange/black/green lines in the right y-axis are sample-wise improvement(↑)/bias(↓)/gain(↑), and the blue bar in the left y-axis is accuracy(↑). 
}
\label{figtradeoff}
\end{figure}

To measure the trade-off between class-dependent bias and accuracy, we take the PTBXL form as our experiment dataset and train the policy network through our framework.
We then introduce different percentages of no-augmentation probability (NoAug percentage) into the trained policy and study the performance change at different NoAug percentages.

Figure \ref{figtradeoff} illustrates that accuracy and sample-wise metrics exhibit optimal performance at varying NoAug percentages. For instance, accuracy(↑) and sample-wise gains(↑) achieve their respective peaks at 50\% and 60\% no-augmentation percentages. Consequently, the application of data augmentation entails a trade-off between accuracy and sample-wise metrics. Thus, when striving to maximize overall performance, it becomes crucial to acknowledge that the implementation of data augmentation may lead to an increase in class-dependent bias, which may deviate from our objective.

\begin{figure}[htbp]
\centerline{\includegraphics[width=80mm,scale=1.0]{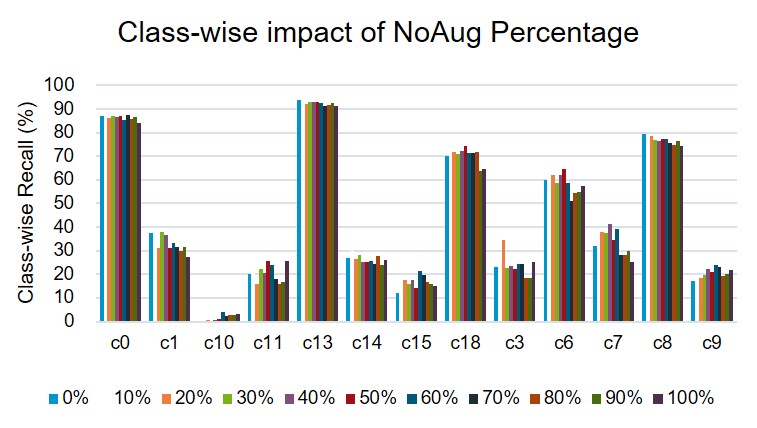}}
\vspace{-0.4cm}
\caption{Class-wise recall changes in different NoAug percentages. The x-axis is the different class index (c0 to c18, remove class with 0\% class-wise recall) in the PTBXL form. Bars in each class index are no-augmentation percentages from 0 (\%) to 100 (\%), and the y-axis is the recall of each class.}
\label{figclass}
\end{figure}

Next, we take a closer look at the relationship between NoAug percentages and class-wise recall to study the performance change of each class.
Figure \ref{figclass} shows that classes with higher recall (e.g., c0, c13, c18) tend to prefer a lower NoAug percentage, while classes with lower recall (e.g., c10, c11, c9) prefer using original samples. This positive trend between class-wise recall and augmentation percentage indirectly explains the design of our Class-dependent Regulation module.

\subsection{Results for External Evaluations}

We evaluate baseline, representative and proposed methods using overall and sample-wise metrics.
Additionally, we select 1d-ResNet as our backbone model to evaluate our proposed CAAP and representative methods.
For all the following experiments, the "\textbf{value}" and "\underline{value}" mean the best and second performance.
Also, the "$-$" represents this method that cannot be evaluated by this metric.


\begin{table*}[htbp]
\caption{Overall Performance on experimental datasets.}
\begin{center}
\resizebox{0.9\textwidth}{55pt}{
\begin{tabular}{l|c|c|c|c|c|c|c|c|c|c|c|c|c|c}
\hline
\textbf{Labelgroup} & \multicolumn{2}{c|}{\textbf{PTBXL sup}}& \multicolumn{2}{c|}{\textbf{PTBXL sub}}& \multicolumn{2}{c|}{\textbf{PTBXL diag}} & \multicolumn{2}{c|}{\textbf{PTBXL rhythm}}& \multicolumn{2}{c|}{\textbf{PTBXL form}}& \multicolumn{2}{c|}{\textbf{Chapman}}& \multicolumn{2}{c}{\textbf{CPSC}} \\
\hline
\textbf{Metric (\%)} & \textbf{ACC}& \textbf{Recall}& \textbf{ACC}& \textbf{Recall}& \textbf{ACC}& \textbf{Recall}& \textbf{ACC}& \textbf{Recall}& \textbf{ACC}& \textbf{Recall}& \textbf{ACC}& \textbf{Recall}& \textbf{ACC}& \textbf{Recall} \\
\midrule
\textbf{NOAUG} & 78.7 & 65.4 & 73.1 & 36.1 & 72.2 & 24.1 & 91.1 & 45.6 & 66.4 & 26.8 & 94.0 & \textbf{49.3} & 84.2 & 79.9 \\
\hline
\textbf{RR permutation} & 79.4 & 66.7 & 73.6 & 36.4 & 72.7 & \underline{25} & \underline{91.4} & 46.0 & 67.2 & \underline{29.0} & 94.0 & 47.3 & 84.2 & 79.3\\
\textbf{QRS resample} & 79.5 & 66.6 & 73.8 & 37.1 & \underline{73.0} & 24.3 & \underline{91.4} & 46.9 & 67.7 & 28.8 & 93.9 & \underline{48.7} & 83.5 & 78.4\\
\hline
\textbf{MODAL} & 79.0 & 66.8 & 73.7 & 36 & 72.3 & 23.7 & 91.2 & 45.8 & 67.4 & 28.7 & 93.9 & 46.9 & \underline{84.6} & \textbf{80.6}\\
\textbf{RandAugment} & \underline{79.7} & 66.8 & \underline{74.1} & \underline{36.9} & 72.6 & 23.5 & \textbf{91.8} & \underline{47.1} & 67.3 & 28 & \underline{94.2} & 44.8 & 84.5 & \underline{80.2}\\
\hline
\textbf{ADDA} & 78.8 & 64.6 & 73.5 & 34.5 & 72.1 & 20.5 & 90.9 & 40.5 & 65.5 & 23.8 & 92.5 & 43.3 & 82.5 & 77.7\\
\textbf{CADDA} & 77.5 & 61.3 & 71.2 & 30.4 & 70.9 & 20.1 & 90.2 & 36.0 & 65.1 & 23.4 & 91.9 & 42.5 & 80.0 & 73.9\\
\textbf{AdaAug} & 79.6 & \underline{66.9} & 73.9 & 35.7 & 72.8 & 22.9 & \textbf{91.8} & 46.5 & \underline{67.8} & 28.1 & 93.5 & 43.5 & 84.1 & 78.9\\
\hline
\textbf{CAAP} & \textbf{80.0} & \textbf{69.3} & \textbf{74.5} & \textbf{39.8} & \textbf{73.2} & \textbf{25.5} & 91.3 & \textbf{50.0} & \textbf{68.6} & \textbf{31.7} & \textbf{94.5} & 46.3 & \textbf{85.2} & \textbf{80.6}\\
\hline
\end{tabular}}
\label{tab_acc1}
\end{center}%
\end{table*}

\textit{Overall Performance Results}: In Table \ref{tab_acc1}, we evaluate the accuracy (ACC) and macro recall (Recall) of each method.
Table \ref{tab_acc1} shows that most data augmentation methods, except CADDA, improve the model's accuracy compared to NOAUG.
However, in some datasets with high baseline performance (e.g., Chapman), augmentation methods do not consistently enhance accuracy and macro recall.
Among the representative methods, AdaAug and RandAugment perform better.
Although CADDA is the first class-wise ADA method, its worsened performance shows that its class-wise method cannot find the optimal policy for transformations.
The representative ADA methods receive less performance improvement when facing more complex tasks, resulting in lower macro recall than the model without data augmentation in the PTBXL diag dataset.

Our CAAP method demonstrates superior accuracy and macro recall across various datasets, exhibiting consistent performance improvement regardless of class numbers. Notably, on the PTBXL diagnostic dataset, our class adaptive framework outperforms representative methods as the class number increases. Additionally, the CAAP method showcases substantial accuracy and macro recall improvement (2.2\% and 4.9\% compared to NOAUG) on the PTBXL form dataset, owing to the suitability of our Information Region Adaption module in preserving ECG waveform properties.
It is worth mentioning that the model without data augmentation achieves the highest recall among all methods in the Chapman dataset. We attribute this difference to the limited size and extreme class distribution of the dataset. However, the CAAP method still demonstrates superior accuracy improvement compared to representative methods in the Chapman dataset.


\begin{table*}[htbp]
\caption{Class-dependent Bias on experimental datasets.}
\begin{center}
\resizebox{\textwidth}{50pt}{
\begin{tabular}{l|c|c|c|c|c|c|c|c|c|c|c|c|c|c}
\hline
\textbf{Labelgroup} & \multicolumn{2}{c|}{\textbf{PTBXL sup}}& \multicolumn{2}{c|}{\textbf{PTBXL sub}}& \multicolumn{2}{c|}{\textbf{PTBXL diag}} & \multicolumn{2}{c|}{\textbf{PTBXL rhythm}}& \multicolumn{2}{c|}{\textbf{PTBXL form}}& \multicolumn{2}{c|}{\textbf{Chapman}}& \multicolumn{2}{c}{\textbf{CPSC}}\\
\hline
\textbf{Metric (\%)} & \textbf{Swise}& \textbf{Swise}& \textbf{Swise}& \textbf{Swise}& \textbf{Swise}& \textbf{Swise} & \textbf{Swise}& \textbf{Swise}& \textbf{Swise}& \textbf{Swise}& \textbf{Swise}& \textbf{Swise}& \textbf{Swise}& \textbf{Swise}\\
 & \textbf{improve $\uparrow$}& \textbf{bias $\downarrow$}& \textbf{improve $\uparrow$}& \textbf{bias $\downarrow$}& \textbf{improve $\uparrow$}& \textbf{bias $\downarrow$}& \textbf{improve $\uparrow$}& \textbf{bias $\downarrow$}& \textbf{improve $\uparrow$}& \textbf{bias $\downarrow$}& \textbf{improve $\uparrow$}& \textbf{bias $\downarrow$}& \textbf{improve $\uparrow$}& \textbf{bias $\downarrow$} \\
\midrule
\textbf{AdaAug} & \underline{6.8} & \underline{5.0} & 4.9 & 5.9 & 4.0 & 3.7 & 6.1 & \underline{4.6} & 4.0 & 7.3 & 1.6 & 8.4 & \underline{5.4} & \underline{5.3}\\
 &(gain)& 1.8 &(gain)& -1.0 &(gain)& 0.3 &(gain)& 1.5 &(gain)& -3.3 &(gain)& -6.8 &(gain)& 0.1 \\
\hline
\textbf{RandAugment} & 6.5 & 5.1 & \underline{5.6} & \underline{5.7} & \underline{4.3} & \underline{3.3} & \underline{6.4} & 5.2 & \underline{5.5} & \underline{5.7} & \textbf{2.2} & \underline{7.6} & \underline{5.4} & 5.4\\
 &(gain)& 1.4 &(gain)& -0.1 &(gain)& 1.0 &(gain)& 1.2 &(gain)& -0.2 &(gain)& -5.4 &(gain)& 0 \\
\hline
\textbf{CAAP} & \textbf{8.0} & \textbf{4.3} & \textbf{7.0} & \textbf{4.1} & \textbf{5.9} & \textbf{2.6} & \textbf{8.5} & \textbf{3.3} & \textbf{7.5} & \textbf{4.6} & \underline{2.1} & \textbf{5.8} & \textbf{5.9} & \textbf{4.4}\\
 &(gain)& \textbf{3.7} &(gain)& \textbf{2.9} &(gain)& \textbf{3.3} &(gain)& \textbf{5.2} &(gain)& \textbf{2.9} &(gain)& \textbf{-3.7} &(gain)& \textbf{1.5} \\
\hline
\multicolumn{15}{l}{Note: $(gain)$ is sample-wise gain of each result, where $gain = improve - bias$.}
\end{tabular}
}
\label{tab_swise1}
\end{center}
\end{table*}

\begin{table*}[htbp]
\caption{Comparison of data augmentation methods in different backbones.}
\begin{center}
\makebox[\textwidth]{
\resizebox{0.9\textwidth}{150pt}{
\begin{tabular}{c|l||c|c|c||c|c|c||c|c|c}
\hline
\multicolumn{2}{c||}{\textbf{Metric (\%)}} & \multicolumn{3}{c||}{\textbf{ACC}} & \multicolumn{3}{c||}{\textbf{Macro Recall}} & \multicolumn{3}{c}{\textbf{Swise gain}} \\
\hline
\multicolumn{2}{c||}{\textbf{Backbone}} & Resnet & FCN & LSTM & Resnet & FCN & LSTM & Resnet & FCN & LSTM \\
\midrule
\multirow{4}{*}{PTBXL sup} 
&\textbf{NOAUG} & 78.7 & 78.9 & 78.5 & 65.4 & \underline{66.5} & 65.6 & - & - & - \\
\cline{2-11}
&\textbf{AdaAug} & 79.6 & \underline{79.4} & \textbf{79.5} & \underline{66.9} & 66.3 & \textbf{67.6} & 1.8 & \underline{-0.2} & \underline{2.1} \\
&\textbf{RandAugment} & \underline{79.7} & 79.2 & \underline{79.4} & 66.8 & 65.8 & 67.0 & \underline{1.4} & -0.8 & \textbf{2.6} \\
\cline{2-11}
&\textbf{CAAP} & \textbf{80.0} & \textbf{79.6} & 79.0 & \textbf{69.3} & \textbf{68.4} & \underline{67.1} & \textbf{3.7} & \textbf{1.9} & 1.6 \\
\hline
\hline
\multirow{4}{*}{PTBXL sub} 
&\textbf{NOAUG} & 73.1 & 73.4 & 72.7 & 36.1 & 34.1 & 34.2 & - & - & - \\
\cline{2-11}
&\textbf{AdaAug} & 73.9 & 73.9 & 73.2 & 35.7 & 34.0 & 34.6 & -1.0 & -0.4 & 1.1 \\
&\textbf{RandAugment} & \underline{74.1} & \underline{74.1} & \textbf{73.9} & \underline{36.9} & \underline{34.9} & \underline{36.7} & \underline{-0.1} & \underline{0.2} & \textbf{2.4} \\
\cline{2-11}
&\textbf{CAAP} & \textbf{74.5} & \textbf{74.3} & \underline{73.4} & \textbf{39.8} & \textbf{36.1} & \textbf{37.6} & \textbf{2.9} & \textbf{1.0} & \underline{1.9} \\
\hline
\hline
\multirow{4}{*}{PTBXL diag} 
&\textbf{NOAUG} & 72.2 & 72.1 & 71.5 & \underline{24.1} & 22.5 & 21.6 & - & - & - \\
\cline{2-11}
&\textbf{AdaAug} & \underline{72.8} & \underline{72.5} & 71.5 & 20.5 & 20.8 & 21.6 & 0.3 & \underline{-1.7} & \underline{0.9} \\
&\textbf{RandAugment} & 72.6 & 72.3 & \underline{72.3} & 23.5 & \underline{23.1} & \underline{22.0} & \underline{1.0} & -1.9 & -1.2 \\
\cline{2-11}
&\textbf{CAAP} & \textbf{73.2} & \textbf{73.0} & \textbf{72.5} & \textbf{25.5} & \textbf{23.2} & \textbf{25.1} & \textbf{3.3} & \textbf{1.0} & \textbf{1.5} \\
\hline
\hline
\multirow{4}{*}{PTBXL rhythm} 
& \textbf{NOAUG} & 91.1 & 90.6 & 91.5 & 45.6 & 44.5 & \textbf{44.3} & - & - & - \\
\cline{2-11}
&\textbf{AdaAug} & \textbf{91.8} & \underline{90.7} & 91.7 & 46.5 & 40.7 & 41.5 & \underline{1.5} & -3.4 & -3.1 \\
&\textbf{RandAugment} & \textbf{91.8} & \textbf{90.9} & \textbf{92.2} & \underline{47.1} & \underline{43.3} & 42.5 & 1.2 & \underline{-1.4} & \underline{-2.0} \\
\cline{2-11}
&\textbf{CAAP} & \underline{91.3} & \underline{90.7} & \underline{92.0} & \textbf{50.0} & \textbf{44.5} & \underline{44.1} & \textbf{5.2} & \textbf{-0.4} & \textbf{0.1} \\
\hline
\hline
\multirow{4}{*}{PTBXL form} 
& \textbf{NOAUG} & 66.4 & 66.1 & 64.1 & 26.8 & 24.9 & 23.1 & - & - & - \\
\cline{2-11}
&\textbf{AdaAug} & \underline{67.8} & 66.2 & 65.5 & \underline{28.1} & 23.2 & 24.5 & -3.3 & -1.8 & 1.3 \\
&\textbf{RandAugment} & 67.3 & \underline{66.5} & \textbf{66.1} & 28.0 & \underline{24.7} & \underline{24.7} & \underline{-0.2} & \underline{-0.3} & \underline{1.4} \\
\cline{2-11}
&\textbf{CAAP} & \textbf{68.6} & \textbf{67.4} & \underline{65.6} & \textbf{31.7} & \textbf{27.0} & \textbf{26.4} & \textbf{2.9} & \textbf{2.2} & \textbf{3.5} \\
\hline
\hline
\multirow{4}{*}{Chapman} 
&\textbf{NOAUG} & 94.0 & 92.0 & \underline{93.4} & \textbf{49.3} & \textbf{45.6} & \textbf{44.8} & - & - & - \\
\cline{2-11}
&\textbf{AdaAug} & 93.5 & \underline{92.1} & 92.2 & 43.5 & 43.0 & 40.0 & -6.8 & \underline{-3.3} & -5.6 \\
&\textbf{RandAugment} & \underline{94.2} & \textbf{92.2} & \textbf{94.1} & 44.8 & 42.7 & \underline{43.1} & \underline{-5.4} & -3.5 & \textbf{-2.4} \\
\cline{2-11}
&\textbf{CAAP} & \textbf{94.5} & \textbf{92.2} & 93.0 & \underline{46.3} & \underline{43.2} & 42.5 & \textbf{-3.7} & \textbf{-3.1} & \underline{-2.9} \\
\hline
\hline
\multirow{4}{*}{CPSC} 
&\textbf{NOAUG} & 84.2 & 81.4 & 83.5 & 79.9 & 76.3 & 78.8 & - & - & - \\
\cline{2-11}
&\textbf{AdaAug} & 84.1 & \underline{82.2} & 84.4 & 78.9 & \underline{77.0} & 80.0 & 0.1 & 0.8 & 1.1 \\
&\textbf{RandAugment} & \underline{84.5} & \underline{82.2} & \underline{85.0} & \underline{80.2} & \underline{77.0} & \underline{80.5} & \underline{0.0} & \underline{0.9} & \underline{1.3} \\
\cline{2-11}
&\textbf{CAAP} & \textbf{85.2} & \textbf{82.5} & \textbf{85.3} & \textbf{80.6} & \textbf{77.7} & \textbf{81.3} & \textbf{1.5} & \textbf{1.5} & \textbf{2.4} \\
\hline
\end{tabular}}}
\label{tab_backbone}
\end{center}%
\end{table*}

\textit{Class-dependent Bias Results}: We evaluate the performance of our proposed CAAP method for reducing class-dependent bias and compare it to two competitive methods, AdaAug and RandAugment. We measure the macro sample-wise class bias (Swise bias), improvement (Swise improve) and gain (sample-wise gain) of each method. Specifically, our goal is to minimize the Swise bias value while maximizing the Swise improvement value. Additionally, the sample-wise gain represents the difference between improvement and bias. A higher gain value indicates superior performance for each class.

Table \ref{tab_swise1} demonstrates that competitive methods exhibit equal or higher sample-wise bias compared to sample-wise improvement (zero or negative sample-wise gain) in half of the datasets, indicating their limited ability to address class-dependent bias. In contrast, our CAAP method achieves positive sample-wise gain and outperforms competitive methods on both Swise improvement and bias in most datasets. The experimental results indicate that our class-wise augmentation policy searching process can learn less biased policies compared to competitive methods. Additionally, the CAAP method achieves a superior macro sample-wise gain of up to 3.7\% compared to competitive methods, demonstrating the advantage of our Class-dependent Regulation module. Furthermore, in datasets with more classes, the CAAP method reaches lower sample-wise bias and better sample-wise improvement, demonstrating the advantages of our class adaptive policy. Even in the Chapman dataset, where all data augmentation methods led to worse sample-wise performance, our proposed method still achieved the lowest sample-wise bias.


Our CAAP method achieves superior overall performance and class-dependent bias for most datasets, especially those with waveform properties or larger class numbers. These results verify the effectiveness of our proposed framework and modules in addressing reducing class-dependent bias and enhancing the model's generalization capability. Notably, even in datasets that may not be ideal for conventional data augmentation approaches, such as the Chapman dataset, our CAAP method still provides accuracy improvements.

\subsection{Results in Different Backbones}

We assess the performance of our proposed CAAP method on three distinct backbone models, namely 1d-ResNet, FCN and LSTM, and conduct a comparative analysis with AdaAug and RandAugment.
Table \ref{tab_backbone} shows the accuracy (ACC), macro recall (Macro Recall) and sample-wise gain (Swise gain) for each method.
As shown in Table \ref{tab_backbone}, the experiment results demonstrate that augmentation methods improve the model's overall performance compared to NOAUG in most datasets, regardless of the backbone used.

CAAP consistently outperforms the competitive methods, achieving superior accuracy and macro recall in most datasets while preserving positive sample-wise gains. Conversely, competitive methods yield negative sample-wise gains in half of the dataset and backbone combinations. In summary, our proposed CAAP method effectively enhances the overall and class-dependent performance of different backbone models and outperforms competitive methods in most datasets. Furthermore, the variation in improvement across different backbones indicates that our method achieves more significant overall enhancements with CNN-based backbone models.

\subsection{Results in Other Time-Series Datasets}

To evaluate the effectiveness of the proposed methods in different domains, we perform additional evaluations on other time series datasets: Human Activity Recognition (HAR) and Electroencephalography (EEG). We select the WISDM dataset \cite{wisdm} and EFDX dataset \cite{edfx} \cite{edfx_physioNet} as our experimental datasets in the HAR and EEG domains, respectively. Since these datasets have different properties than ECG datasets, we first conduct a grid search on the model without data augmentation to determine the appropriate hyper-parameters. Subsequently, we evaluate our proposed CAAP and competitive methods on different datasets. The statistic of other time series datasets is shown in Table \ref{tab_data_ts}, including the number of records (Records), the number of classes (Class nums), the sampling rate (Sfreq (Hz)), the real-time length of the sample (Length) and the number of channels (Channel nums).

\begin{table}[htbp]
\caption{The characteristics of other time series datasets after pre-processing.}
\begin{center}
\makebox[\textwidth]{
\resizebox{0.6\textwidth}{15pt}{
\begin{tabular}{|c|c|c|c|c|c|}
\hline
\textbf{Dataset} & \textbf{Records}& \textbf{Class nums}& \textbf{Sfreq (Hz)}& \textbf{Length}& \textbf{Channel nums} \\
\midrule
\textbf{WISDM} & 23995 & 18 & 20 & 10 s & 3 \\
\hline
\textbf{EDFX} & 85188 & 5 & 100 & 30 s & 2 \\
\hline
\end{tabular}}}
\label{tab_data_ts}
\end{center}
\end{table}

We evaluate the accuracy and sample-wise metrics on the other time series datasets to assess the effectiveness of the proposed methods in different time series datasets. The competitive methods selected for comparison are AdaAug and RandAugment due to their advanced performance in external experiments.

\begin{table}[htbp]
\caption{Comparison of data augmentation methods on other time series datasets.}
\begin{center}
\makebox[\textwidth]{
\resizebox{0.9\textwidth}{45pt}{
\begin{tabular}{|c|c||c|c||c|c|c|}
\hline
\multicolumn{2}{|c||}{\textbf{Metric (\%)}} & \textbf{ACC $\uparrow$} & \textbf{Macro Recall $\uparrow$} & \textbf{Swise improve $\uparrow$}& \textbf{Swise bias $\downarrow$} & \textbf{Swise gain $\uparrow$} \\
\midrule
\multirow{4}{*}{WISDM} 
&\textbf{NOAUG} & \underline{86.8} & \underline{86.8} & - & - & - \\
\cline{2-7}
&\textbf{AdaAug} & 86.1 & 86.1 & 4.0 & 4.8 & -0.8 \\
&\textbf{RandAugment} & 85.9 & 85.9 & \textbf{4.4} & \underline{4.5} & \underline{-0.1} \\
\cline{2-7}
&\textbf{CAAP} & \textbf{87.4} & \textbf{87.4} & \textbf{4.4} & \textbf{3.8} & \textbf{0.6} \\
\hline
\hline
\multirow{4}{*}{EDFX} 
&\textbf{NOAUG} & 81.5 & 73.7 & - & - & - \\
\cline{2-7}
&\textbf{AdaAug} & 82.5 & 75.0 & 6.1 & 4.8 & 1.3 \\
&\textbf{RandAugment} & \textbf{83.0} & \underline{75.8} & \underline{6.6} & \underline{4.5} & \underline{2.1} \\
\cline{2-7}
&\textbf{CAAP} & \underline{82.8} & \textbf{76.2} & \textbf{6.9} & \textbf{4.4} & \textbf{2.5} \\
\hline
\end{tabular}}}
\label{tab_ts}
\end{center}
\end{table}

In Table \ref{tab_ts}, we evaluate the accuracy (ACC), macro recall (Macro Recall) and sample-wise metrics (Swise improve, Swise bias and Swise gain) of each method on the WISDM and EDFX datasets. The experimental results indicate that competitive methods may not consistently enhance accuracy in other time series datasets. In contrast, our CAAP method consistently achieves superior or similar overall performance across these datasets. For the WISDM dataset, while competitive methods can have adverse effects compared to models without data augmentation, our CAAP method maintains accuracy improvements and reduces class-dependent bias. It's important to note that macro recall equals accuracy on the WISDM dataset, where all classes are perfectly balanced. Regarding the EDFX dataset, our CAAP method achieves similar accuracy and lower class-dependent bias than RandAugment. In summary, our additional evaluations underscore the effectiveness of the proposed CAAP method across various time series datasets, including HAR and EEG datasets. 


\subsection{Results for Internal Evaluations}

\begin{table*}[htbp]
\caption{Overall performance for removing modules on experimental datasets.}
\begin{center}
\resizebox{0.9\textwidth}{35pt}{
\begin{tabular}{l|c|c|c|c|c|c|c|c|c|c|c|c|c|c}
\hline
\textbf{Labelgroup} & \multicolumn{2}{c|}{\textbf{PTBXL sup}}& \multicolumn{2}{c|}{\textbf{PTBXL sub}}& \multicolumn{2}{c|}{\textbf{PTBXL diag}} & \multicolumn{2}{c|}{\textbf{PTBXL rhythm}}& \multicolumn{2}{c|}{\textbf{PTBXL form}}& \multicolumn{2}{c|}{\textbf{Chapman}}& \multicolumn{2}{c}{\textbf{CPSC}}\\
\hline
\textbf{Metric (\%)} & \textbf{ACC}& \textbf{Recall}& \textbf{ACC}& \textbf{Recall}& \textbf{ACC}& \textbf{Recall} & \textbf{ACC}& \textbf{Recall}& \textbf{ACC}& \textbf{Recall}& \textbf{ACC}& \textbf{Recall}& \textbf{ACC}& \textbf{Recall}\\
\midrule
\textbf{CAAP-p} & \textbf{80.4} & \textbf{68.8} & \textbf{74.8} & \textbf{38.1} & \textbf{73.4} & \textbf{23.9} & \textbf{91.7} & 47.4 & \textbf{68.5} & 30.5 & \textbf{94.4} & \textbf{46.4} & 84.8 & 80.6\\
\hline
\textbf{CAAP-p (-diff)} & 80.1 & 68.5 & 74.2 & 37.5 & 73.1 & 22.9 & 91.6 & 47.3 & 68.2 & 29.0 & 94.2 & 45.4 & 85.0 & 80.5\\
\textbf{CAAP-p (-info\_reg)} & 79.9 & 67.3 & 74.2 & 36.2 & 73.0 & 22.8 & \textbf{91.7} & 45.5 & 67.0 & 25.1 & 93.6 & 43.9 & 84.4 & 79.9\\
\textbf{CAAP-p (-bal)} & 79.9 & 67.5 & 74.6 & 37.7 & \textbf{73.4} & \textbf{23.9} & \textbf{91.7} & \textbf{48.8} & 68.4 & \textbf{30.6} & \textbf{94.4} & 45.9 & \textbf{85.3} & \textbf{81.2}\\
\hline
\multicolumn{15}{l}{Note: The CAAP-p stands for CAAP without the Class-dependent Regulation module.}
\end{tabular}}
\label{tab_internal1}
\end{center}
\end{table*}

\textit{Effects of Different Modules}:
First, we evaluate the overall performance of our CAAP framework by removing different modules and measuring the accuracy and macro recall in the 1d-Resnet backbone. The origin method we used is CAAP without the Class-dependent Regulation module (CAAP-p) because we focus on evaluating overall performance in this experiment.
There are three modules we will evaluate in this experiment: Difficult policy loss (diff), info region module (info\_reg) and balance sampler (bal).
In Table \ref{tab_internal1}, their removal resulted in significant performance drops in most datasets, which shows that all modules contribute to our framework.
The significant performance degradation of CAAP-p (-info\_reg) indicates that considering informative regions is helpful in the ECG task.
Notably, the CAAP-p (-info\_reg) caused a 5\% macro recall and a 1.5\% accuracy drop in the PTBXL form dataset, demonstrating its crucial role in maintaining important ECG form features for model learning.

\begin{table*}[htbp]
\caption{Class-dependent Bias for removing modules on experimental datasets.}
\begin{center}
\resizebox{\textwidth}{60pt}{
\begin{tabular}{l|c|c|c|c|c|c|c|c|c|c|c|c|c|c}
\hline
\textbf{Labelgroup} & \multicolumn{2}{c|}{\textbf{PTBXL sup}}& \multicolumn{2}{c|}{\textbf{PTBXL sub}}& \multicolumn{2}{c|}{\textbf{PTBXL diag}} & \multicolumn{2}{c|}{\textbf{PTBXL rhythm}}& \multicolumn{2}{c|}{\textbf{PTBXL form}}& \multicolumn{2}{c|}{\textbf{Chapman}}& \multicolumn{2}{c}{\textbf{CPSC}}\\
\hline
\textbf{Metric (\%)} & \textbf{Swise}& \textbf{Swise}& \textbf{Swise}& \textbf{Swise}& \textbf{Swise}& \textbf{Swise} & \textbf{Swise}& \textbf{Swise}& \textbf{Swise}& \textbf{Swise}& \textbf{Swise}& \textbf{Swise}& \textbf{Swise}& \textbf{Swise}\\
 & \textbf{improve $\uparrow$}& \textbf{bias $\downarrow$}& \textbf{improve $\uparrow$}& \textbf{bias $\downarrow$}& \textbf{improve $\uparrow$}& \textbf{bias $\downarrow$}& \textbf{improve $\uparrow$}& \textbf{bias $\downarrow$}& \textbf{improve $\uparrow$}& \textbf{bias $\downarrow$}& \textbf{improve $\uparrow$}& \textbf{bias $\downarrow$}& \textbf{improve $\uparrow$}& \textbf{bias $\downarrow$} \\
\midrule
\textbf{CAAP-p} & \textbf{7.6} & \textbf{4.3} & 6.0 & \textbf{4.1} & \textbf{4.9} & \textbf{3.0} & \textbf{7.3} & \textbf{4.0} & 6.6 & 4.9 & 1.7 & \textbf{5.4} & 5.6 & 4.2\\
 &(gain)& \textbf{3.3} &(gain)& \textbf{1.9} &(gain)& \textbf{1.9} &(gain)& \textbf{3.3} &(gain)& 1.7 &(gain)& \textbf{-3.7} &(gain)& 1.4 \\
 \hline
\textbf{CAAP-p (-diff)} & 7.4 & 4.5 & 6.0 & 5.7 & 4.1 & 3.3 & 6.9 & 4.7 & 5.9 & 5.7 & 1.6 & 6.4 & 5.1 & 3.8\\
 &(gain)& 2.9 &(gain)& 0.3 &(gain)& 0.8 &(gain)& 2.2 &(gain)& 0.2 &(gain)& -4.8 &(gain)& 1.3 \\
\hline
\textbf{CAAP-p (-info\_reg)} & 6.8 & 5.0 & 5.7 & 6.1 & 3.8 & 3.5 & 7.0 & 6.4 & 4.4 & 8.1 & 1.7 & 8.1 & 5.5 & 4.7\\
 &(gain)& 1.8 &(gain)& -0.4 &(gain)& 0.3 &(gain)& 0.6 &(gain)& -3.7 &(gain)& -6.4 &(gain)& 0.8 \\
\hline
\textbf{CAAP-p (-bal)} & 7.5 & 4.4 & \textbf{6.1} & 4.8 & 4.8 & 3.0 & 7.0 & 4.2 & \textbf{7.1} & \textbf{4.7} & \textbf{2.0} & 6.3 & \textbf{6.0} & \textbf{4.0}\\
 &(gain)& 3.1 &(gain)& 1.3 &(gain)& 1.8 &(gain)& 2.8 &(gain)& \textbf{2.4} &(gain)& -4.3 &(gain)& \textbf{2.0} \\
\hline
\multicolumn{15}{l}{Note: $(gain)$ is sample-wise gain of each result, where $gain = improve - bias$.}
\end{tabular}
}
\label{tab_internal2}
\end{center}
\end{table*}

Second, we evaluate the class-dependent bias of our CAAP framework by removing different modules in the 1d-Resnet backbone.
As shown in Table \ref{tab_internal2}, removing any module resulted in significant sample-wise performance declines across most datasets. This underscores the role of all modules in reducing class-dependent bias within our framework. Moreover, the result also illustrates the significant sample-wise gain degradation of CAAP-p (-info\_reg), further emphasizing the ECG task's informative regions.

Regarding the PTBXL form and CPSC datasets, CAAP-p (-bal) outperforms the full CAAP-p in terms of sample-wise gain, indicating that the standard data sampler is more effective for smaller datasets with uniform class distribution. Despite the balance sampler not consistently leading to superior performance, we retain CAAP with the balance sampler as our primary proposed framework, as it demonstrates better overall and sample-wise performance in most datasets.

\begin{table}[htbp]
\caption{Sample-wise Gain for different reweight methods.}
\begin{center}
\resizebox{0.6\linewidth}{40pt}{
\begin{tabular}{l|c|c|c}
\hline
\textbf{Method} & \textbf{CAAP-p}& \textbf{CAAP (cadd0.5)}& \textbf{CAAP (cadd1.0)} \\
\midrule
\textbf{PTBXL sup} & 3.3 & \textbf{3.7} & \textbf{3.7} \\
\textbf{PTBXL sub} & 1.9 & \textbf{2.9} & \textbf{2.9} \\
\textbf{PTBXL diag} & 1.9 & \textbf{3.3} & 2.9 \\
\textbf{PTBXL rhythm} & 3.3 & \textbf{5.2} & 3.1 \\
\textbf{PTBXL form} & 1.7 & 2.9 & \textbf{3.9} \\
\textbf{Chapman} & \textbf{-3.7} & \textbf{-3.7} & -5.7 \\
\textbf{CPSC} & 1.4 & \textbf{1.5} & 1.1 \\
\hline
\end{tabular} }
\label{tab_rew}
\end{center}
\end{table}

\textit{Comparisons of Different NoAug Reweight Methods}: We evaluate the effectiveness of our Class-dependent Regulation module. Three methods are considered: Without Class-dependent Regulation module (CAAP-p) and Class-dependent Regulation module with $\alpha_{NoAug}$ values of 0.5 and 1.0 (cadd0.5, cadd1.0). The evaluation of sample-wise gain for each reweighting method is presented in Table \ref{tab_rew}. The results indicate that reweighting methods outperform the CAAP-p, indicating the efficacy of NoAug reweighting in mitigating class-dependent bias. Notably, the module with $\alpha_{NoAug}=0.5$ (cadd0.5) demonstrates superior sample-wise metrics in most datasets. This result confirms the appropriateness of using $\alpha_{NoAug}=0.5$ as the reweighting $\alpha_{NoAug}$ value for our Class-dependent Regulation module.

\textit{Case Study for Information Region Adaption}: In this section, we present a comprehensive case study to evaluate the effectiveness of our proposed Information Region Adaptation module in identifying informative regions in ECG signals. This module has been identified as the most significant contributor to our framework through previous internal experiments. By examining information regions in different categories of ECG signals, we aim to demonstrate the module's efficacy.
For our case study, we select the PTBXL sub dataset and utilize the 1d-ResNet backbone model as our experimental dataset and backbone model, respectively. Furthermore, we focus on three target classes for in-depth analysis, which are described as follows:
\begin{itemize}
    \item STTC (class index = 20):
    STTC refers to abnormal waveform changes between the S and T waves or within the T wave.
    \item RAO/RAE (class index = 17):
    RAO/RAE occurs when the right atrium is larger than usual, resulting in higher P waves \cite{rae_ecg}.
    \item AMI (class index = 0):
    AMI is a type of myocardial infarction occurring due to a decrease in blood supply to the anterior wall of the heart.
    It is characterized by an elevation in the J Point (the point between the QRS and ST wave) \cite{ami_ecg}.
\end{itemize}

\begin{figure*}[htbp]
\begin{center}
\subfloat[STTC]{\includegraphics[width=70mm,scale=1.0]{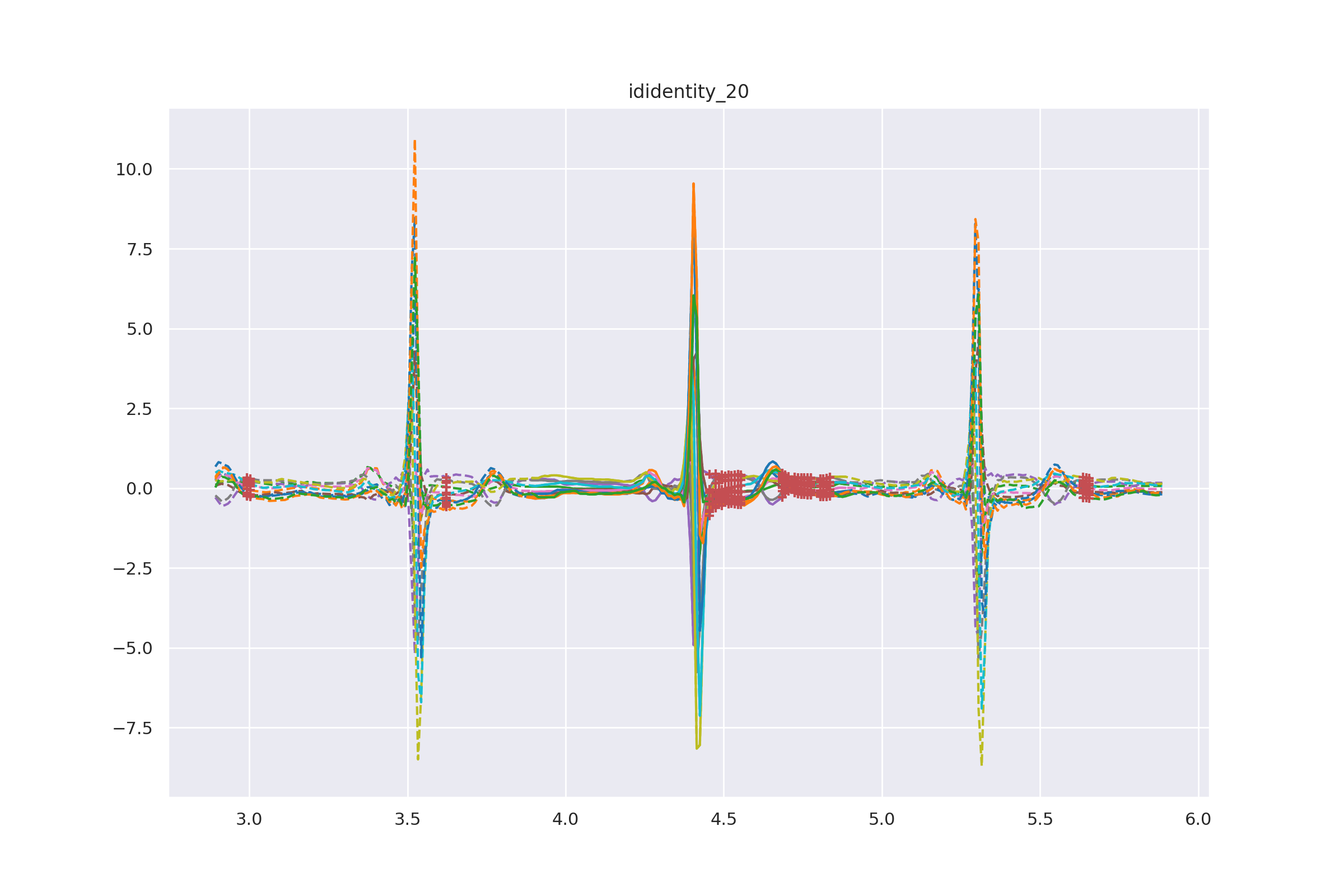}}
\subfloat[RAO/RAE]{\includegraphics[width=70mm,scale=1.0]{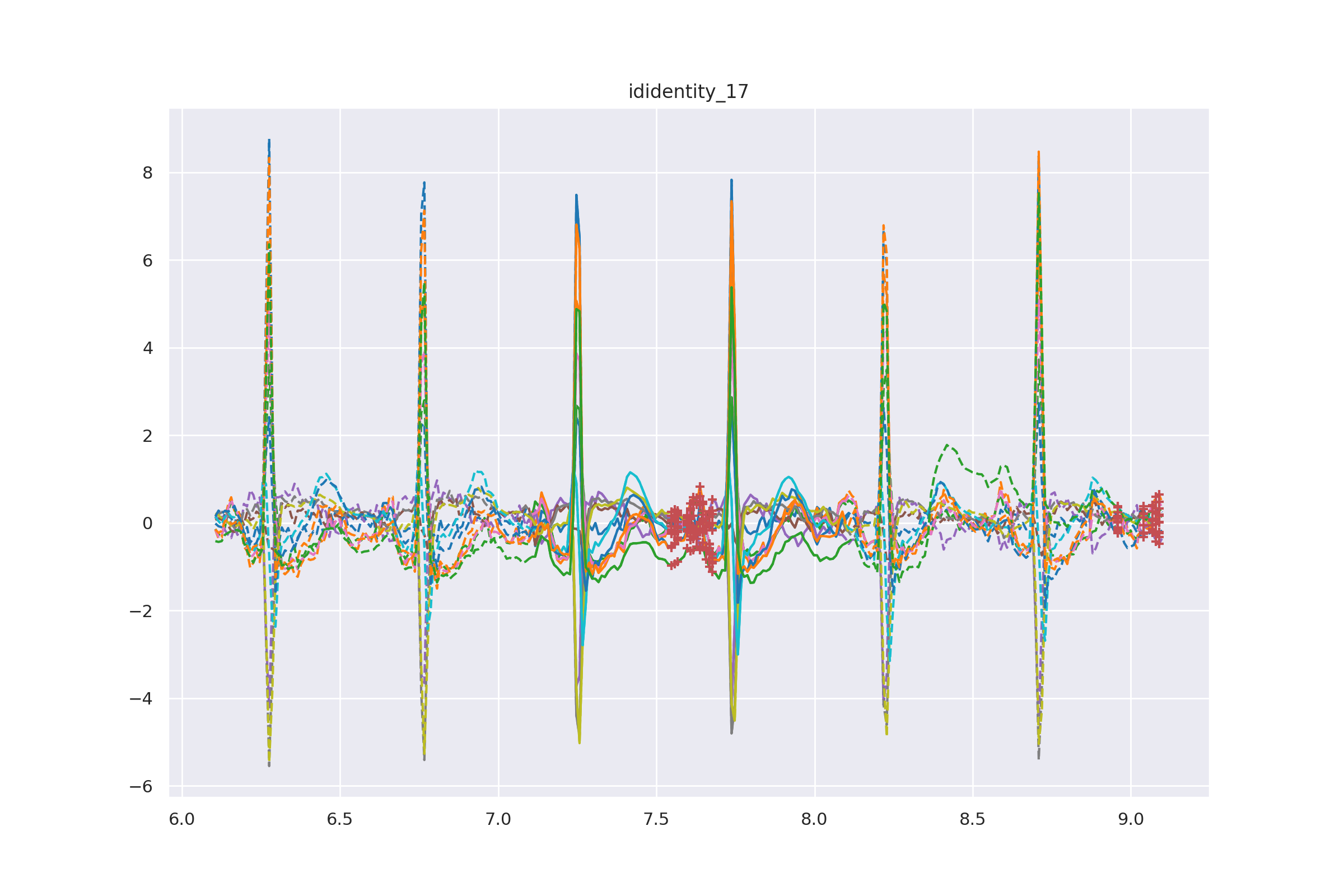}}
\\
\vspace{-0.4cm}
\subfloat[AMI]{\includegraphics[width=70mm,scale=1.0]{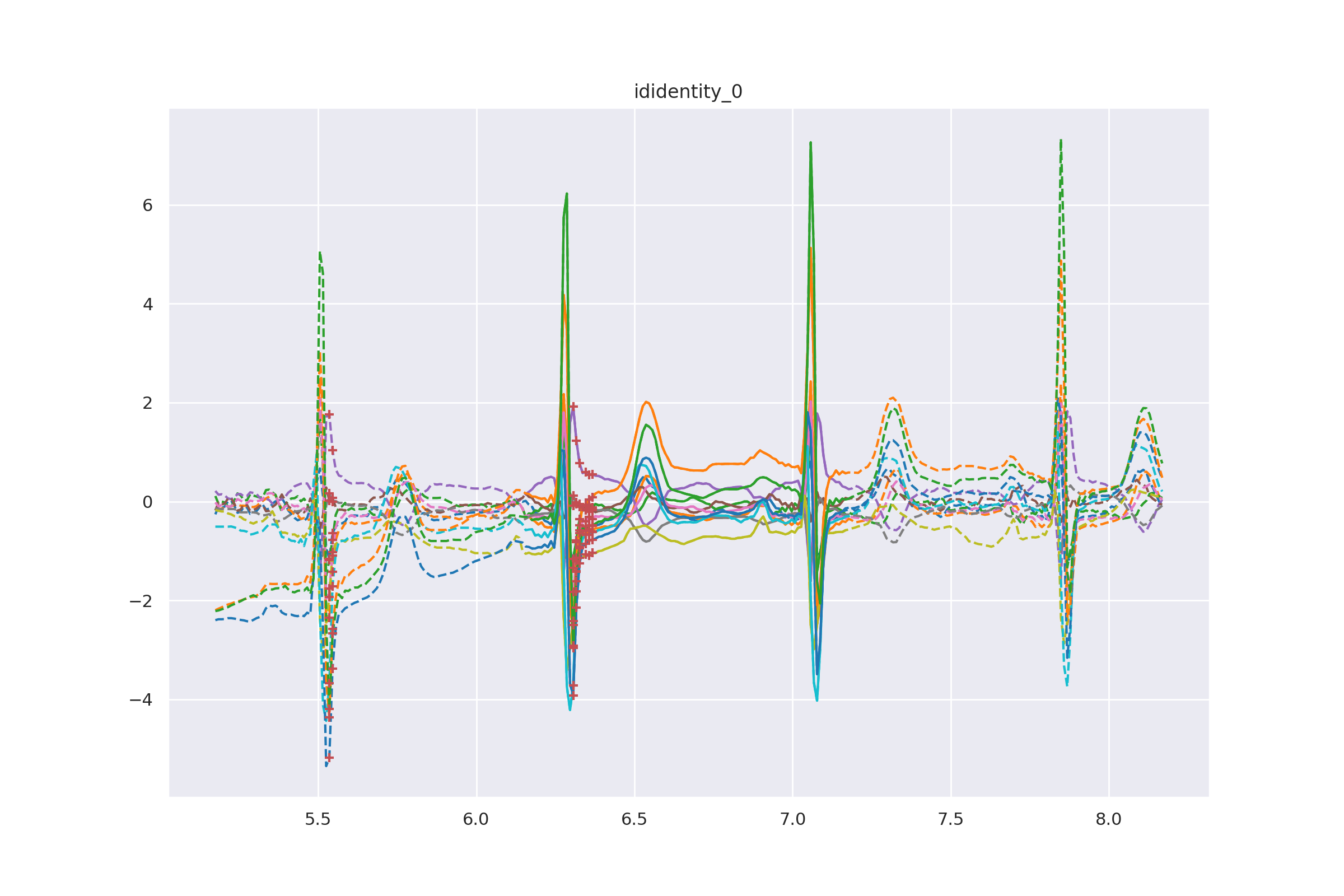}}
\end{center}
\vspace{-0.5cm}
\caption{Visualization of the information regions in different classes.}%
{
The (a), (b) and (c) corresponds to the ECG signals from STTC, RAO/RAE and AMI class.
The colored lines are different leads in the ECG signal, and the red markers ($+$) represent time steps with importance scores exceeding 0.4.
The solid line represents the selected information region, while the dotted line denotes the surrounding ECG signal.
Also, the x-axis represents the time step in seconds, while the y-axis represents the normalized value.
}
\label{fig_case_ecg}
\end{figure*}

As depicted in Figure \ref{fig_case_ecg} (a), the high-importance regions predominantly coincide with the ST and T waves, allowing the identification of delayed ST waves within the ECG signal. Figure \ref{fig_case_ecg} (b) demonstrates that crucial time steps are concentrated in the higher P waves of the ECG signal. In Figure \ref{fig_case_ecg} (c), the most significant regions are observed in the QRS waves and the J Point. Our proposed module focuses on the point following the QRS complex (J Point) and randomly selects a region exhibiting abnormal J Point characteristics. The outcomes presented in Figure \ref{fig_case_ecg} affirm the effectiveness of our Information Region Adaptation module in identifying critical components across different classes. Subsequently, our module selectively protects regions with high-importance scores during augmentation.

\section{Conclusions}\label{conclusion}

This work addressed the challenge of reducing \textit{class-dependent bias} while improving overall performance in Automatic Data Augmentation. This type of bias has not been adequately solved in the existing literature, which unfortunately exerts an unavoidable negative influence on practical applications. To tackle this challenge, we have proposed the novel \textit{Class-dependent Automatic Adaptive Policies (CAAP)} framework, which incorporates three well-designed modules. In particular, the \textit{Class Adaption Policy Network} is designed to learn suitable augmentation sample-wise policies, regardless of the number and distribution of classes. Additionally, the \textit{Class-dependent Regulation} module adjusts no-augmentation probability to mitigate class-dependent bias after the searching phase. As a specialized module designed for class-dependent bias, it has the potential to function independently and collaborate effectively with other advanced methods. The \textit{Information Region Adaptation} module intelligently preserves crucial information that is often overlooked in the field of ADA. This module is particularly advantageous for time series data with waveform characteristics. Furthermore, we have introduced a novel metric to measure the class-dependent bias quantitatively and examined the relationship between accuracy and bias when applying augmentation policies, providing valuable guidance and insights for future studies in this area.

Experimental results on real-world, large-scale ECG datasets demonstrate that the CAAP framework achieved a sample-wise gain increase of up to 3.7\%, accompanied by a 0.8\% increase in accuracy and 2.9\% increase in macro recall compared to the best competitive method. As an effective ADA method, it improves accuracy and macro recall by up to 2.2\% and 4.9\%, respectively, compared to the absence of augmentation. Both external and internal experiments reinforce the conclusion that the CAAP framework effectively reduces class-dependent bias while maintaining superior overall performance. These outcomes establish the reliability of CAAP as an ADA method, addressing the requirements of real-world applications.


\bibliographystyle{unsrt}  
\bibliography{references}  

\end{document}